# Fast, Small and Exact: Infinite-order Language Modelling with Compressed Suffix Trees

**Ehsan Shareghi,**[♭] **Matthias Petri,**[♮] **Gholamreza Haffari**[♭] **and Trevor Cohn**[♮]
[♭] Faculty of Information Technology, Monash University
[♮] Computing and Information Systems, The University of Melbourne
`first.last@monash.edu, initial.last@unimelb.edu.au`

## Abstract

Efficient methods for storing and querying are critical for scaling high-order $m$-gram language models to large corpora. We propose a language model based on *compressed suffix trees*, a representation that is highly compact and can be easily held in memory, while supporting queries needed in computing language model probabilities on-the-fly. We present several optimisations which improve query runtimes up to 2500×, despite only incurring a modest increase in construction time and memory usage. For large corpora and high Markov orders, our method is highly competitive with the state-of-the-art KenLM package. It imposes much lower memory requirements, often by orders of magnitude, and has runtimes that are either similar (for training) or comparable (for querying).

## 1 Introduction

Language models (LMs) are fundamental to many NLP tasks, including machine translation and speech recognition. Statistical LMs are probabilistic models that assign a probability to a sequence of words $w_1^N$, indicating how likely the sequence is in the language. $m$-gram LMs are popular, and prove to be accurate when estimated using large corpora. In these LMs, the probability of $m$-grams are often precomputed and stored explicitly.

Although widely successful, current $m$-gram LM approaches are impractical for learning high-order LMs on large corpora, due to their poor scaling properties in both training and query phases. Prevailing methods (Heafield, 2011; Stolcke et al., 2011) precompute all $m$-gram probabilities, and consequently need to store and access as many as a hundred of billions of $m$-grams for a typical moderate-order LM.

Recent research has attempted to tackle scalability issues through the use of efficient data structures such as tries and hash-tables (Heafield, 2011; Stolcke et al., 2011), lossy compression (Talbot and Osborne, 2007; Levenberg and Osborne, 2009; Guthrie and Hepple, 2010; Pauls and Klein, 2011; Church et al., 2007), compact data structures (Germann et al., 2009; Watanabe et al., 2009; Sorensen and Allauzen, 2011), and distributed computation (Heafield et al., 2013; Brants et al., 2007). Fundamental to all the widely used methods is the precomputation of *all* probabilities, hence they do not provide an adequate trade-off between space and time for high $m$, both during training and querying. Exceptions are Kennington et al. (2012) and Zhang and Vogel (2006), who use a suffix-tree or suffix-array over the text for computing the sufficient statistics on-the-fly.

In our previous work (Shareghi et al., 2015), we extended this line of research using a Compressed Suffix Tree (CST) (Ohlebusch et al., 2010), which provides a considerably more compact searchable means of storing the corpus than an uncompressed suffix array or suffix tree. This approach showed favourable scaling properties with $m$ and had only a modest memory requirement. However, the method only supported Kneser-Ney smoothing, not its modified variant (Chen and Goodman, 1999) which overall performs better and has become the de-facto standard. Additionally, querying was significantly slower than for leading LM toolkits, making the method impractical for widespread use.

In this paper we extend Shareghi et al. (2015) to support modified Kneser-Ney smoothing, and

present new optimisation methods for fast construction and querying.[1] Critical to our approach are:

- Precomputation of several *modified counts*, which would be very expensive to compute at the query time. To orchestrate this, a subset of the CST nodes is selected based on the cost of computing their modified counts (which relates with the branching factor of a node). The precomputed counts are then stored in a compressed data structure supporting efficient memory usage and lookup.

- Re-use of CST nodes within $m$-gram probability computation as a sentence gets scored left-to-right, thus saving many expensive lookups.

Empirical comparison against our earlier work (Shareghi et al., 2015) shows the significance of each of these optimisations. The strengths of our method are apparent when applied to very large training datasets ($\geq 16$ GiB) and for high order models, $m \geq 5$. In this setting, while our approach is more memory efficient than the leading KenLM model, both in the construction (training) and querying phases (testing), we are highly competitive in terms of runtimes of both phases. When memory is a limiting factor at query time, our approach is orders of magnitude faster than the state of the art. Moreover, our method allows for efficient querying with an unlimited Markov order, $m \to \infty$, without resorting to approximations or heuristics.

## 2 Modified Kneser-Ney Language Model

In an $m$-gram language model, the probability of a sentence is decomposed into $\prod_{i=1}^{N} P(w_i|w_{i-m+1}^{i-1})$, where $P(w_i|w_{i-m+1}^{i-1})$ is the conditional probability of the next word given its finite history. Smoothing techniques are employed to deal with sparsity when estimating the parameters of $P(w_i|w_{i-m+1}^{i-1})$. A comprehensive comparison of different smoothing techniques is provided in (Chen and Goodman, 1999). We focus on interpolated Modified Kneser-Ney (MKN) smoothing, which is widely regarded as a state-of-the-art technique and is supported by popular language modelling toolkits, e.g. SRILM (Stolcke, 2002) and KenLM (Heafield, 2011).

[1] https://github.com/eehsan/cstlm

$$P_m(w|u\mathbf{x}) = \frac{[c(u\mathbf{x}w) - \mathbb{D}^m(c(u\mathbf{x}w))]^+}{c(u\mathbf{x})} + \frac{\gamma^m(u\mathbf{x})\bar{P}_{m-1}(w|\mathbf{x})}{c(u\mathbf{x})}$$

$$\bar{P}_k(w|u\mathbf{x}) = \frac{[N_{1+}(\bullet u\mathbf{x}w) - \mathbb{D}^k(N_{1+}(\bullet u\mathbf{x}w))]^+}{N_{1+}(\bullet u\mathbf{x}\bullet)} + \frac{\gamma^k(u\mathbf{x})\bar{P}_{k-1}(w|\mathbf{x})}{N_{1+}(\bullet u\mathbf{x}\bullet)}$$

$$\bar{P}_0(w|\epsilon) = \frac{[N_{1+}(\bullet w) - \mathbb{D}^1(N_{1+}(\bullet w))]^+}{N_{1+}(\bullet\bullet)} + \frac{\gamma(\epsilon)}{N_{1+}(\bullet\bullet)} \times \frac{1}{\sigma}$$

$$\gamma^k(u\mathbf{x}) = \begin{cases} \sum_{j\in\{1,2,3+\}} \mathbb{D}^k(j) N_j(u\mathbf{x}\bullet), & \text{if } k = m \\ \sum_{j\in\{1,2,3+\}} \mathbb{D}^k(j) N'_j(u\mathbf{x}\bullet), & \text{if } k < m \end{cases}$$

$$\mathbb{D}^k(j) = \begin{cases} 0, & \text{if } j = 0 \\ 1 - 2\frac{n_2(k)}{n_1(k)} \cdot \frac{n_1(k)}{n_1(k)+2n_2(k)}, & \text{if } j = 1 \\ 2 - 3\frac{n_3(k)}{n_2(k)} \cdot \frac{n_1(k)}{n_1(k)+2n_2(k)}, & \text{if } j = 2 \\ 3 - 4\frac{n_4(k)}{n_3(k)} \cdot \frac{n_1(k)}{n_1(k)+2n_2(k)}, & \text{if } j \geq 3 \end{cases}$$

$$n_i(k) = \begin{cases} |\{\alpha \text{ s.t. } |\alpha| = k, c(\alpha) = i\}|, & \text{if } k = m \\ |\{\alpha \text{ s.t. } |\alpha| = k, N_{1+}(\bullet\alpha) = i\}|, & \text{if } k < m \end{cases}$$

Figure 1: The quantities and formula needed for modified Kneser-Ney smoothing, where $\mathbf{x}$ is a $k$-gram, $u$ and $w$ are words, and $[a]^+ \stackrel{\text{def}}{=} \max\{0, a\}$. We use $m$ to refer to the order of the language model, and $k \in [1, m]$ to the level of smoothing. The recursion stops at the unigram level $\bar{P}_0(w|\epsilon)$ where the probability is smoothed by the uniform distribution over the vocabulary $\frac{1}{\sigma}$.

MKN is a recursive smoothing technique which uses lower order $k$-gram language models to smooth higher order models. Figure 1 describes the recursive smoothing formula employed in MKN. It is distinguished from Kneser-Ney (KN) smoothing in its use of *adaptive* discount parameters (denoted as $\mathbb{D}^k(j)$ in Figure 1) based on the $k$-gram counts. Importantly, MKN is based on not just $m$-gram frequencies, $c(\mathbf{x})$, but also several *modified counts* based on numbers of unique contexts, namely

$$N_{i+}(\alpha\bullet) = |\{w \text{ s.t. } c(\alpha w) \geq i\}|$$
$$N_{i+}(\bullet\alpha) = |\{w \text{ s.t. } c(w\alpha) \geq i\}|$$
$$N_{i+}(\bullet\alpha\bullet) = |\{w_1 w_2 \text{ s.t. } c(w_1\alpha w_2) \geq i\}|$$
$$N'_{i+}(\alpha\bullet) = |\{w \text{ s.t. } N_{1+}(\bullet\alpha w) \geq i\}| .$$

$N_{i+}(\bullet\alpha)$ and $N_{i+}(\alpha\bullet)$ are the number of words with frequency at least $i$ that come before and after a pattern $\alpha$, respectively. $N_{i+}(\bullet\alpha\bullet)$ is the number of word-pairs with frequency at least $i$ which surround $\mathbf{x}$. $N'_{i+}(\alpha\bullet)$ is the number of words coming after $\alpha$ to form a pattern $\alpha w$ for which the number of unique left contexts is at least $i$; it is specific to MKN and not needed in KN. Table 1 illustrates the different types of quantities required for computing an example 4-gram MKN probability.

| $k$ | num. count | denom. count | $\gamma$ |
|---|---|---|---|
| 4 | $c(\text{Force is strong with})$ | $c(\text{Force is strong})$ | $N_{\{1,2,3+\}}(\text{Force is strong} \bullet)$ |
| 3 | $N_{1+}(\bullet \text{is strong with})$ | $N_{1+}(\bullet \text{is strong})$ | $N'_{\{1,2,3+\}}(\text{is strong} \bullet)$ |
| 2 | $N_{1+}(\bullet \text{strong with})$ | $N_{1+}(\bullet \text{strong})$ | $N'_{\{1,2,3+\}}(\text{strong} \bullet)$ |
| 1 | $N_{1+}(\bullet \text{with})$ | $N_{1+}(\bullet \bullet)$ | $N'_{\{1,2,3+\}}(\epsilon \bullet)$ |

Table 1: The main quantities required for computing $P(\text{with}|\text{Force}, \text{is}, \text{strong})$ under MKN.

Efficient computation of these quantities is challenging with limited memory and time resources, particularly when the order of the language model $m$ is high and/or the training corpus is large. In this paper, we make use of advanced data structures to efficiently obtain the required quantities to answer probabilistic queries as they arrive. Our solution involves precomputing and caching expensive quantities, $N_{1+}(\bullet \alpha \bullet)$, $N_{1+}(\bullet \alpha)$, $N_{\{1,2,3+\}}(\bullet \alpha)$ and $N'_{\{1,2,3+\}}(\alpha \bullet)$, which we will explain in §4. We start in §3 by providing a review of the approach in Shareghi et al. (2015) upon which we base our work.

## 3 KN with Compressed Suffix Trees

### 3.1 Compressed Data Structures

Shareghi et al. (2015) proposed a method for Kneser-Ney (KN) language modelling based on on-the-fly probability computation from a *compressed suffix tree (*CST*)* (Ohlebusch et al., 2010). The CST emulates the functionality of the Suffix Tree (*ST*) (Weiner, 1973) using substantially less space. The suffix tree is a classical search index consisting of a rooted labelled search tree constructed from a text $\mathcal{T}$ of length $n$ drawn from an alphabet of size $\sigma$. Each root to leaf path in the suffix tree corresponds to a suffix of $\mathcal{T}$. The leaves, considered in left-to-right order define the suffix array (*SA*) (Manber and Myers, 1993) such that the suffix $\mathcal{T}[SA[i], n-1]$ is lexicographically smaller than $\mathcal{T}[SA[i+1], n-1]$ for $i \in [0, n-2]$. Searching for a pattern $\alpha$ of length $m$ in $\mathcal{T}$ can be achieved by finding the "highest" node $v$ in the *ST* such that the path from the root to $v$ is prefixed by $\alpha$. All leaf nodes in the subtree starting at $v$ correspond to the locations of $\alpha$ in $\mathcal{T}$. This is translated to finding the specific range $SA[lb, rb]$ such that

$$\mathcal{T}[SA[j], SA[j+m-1]] = \alpha \ \texttt{for} \ j \in [lb, rb]$$

as illustrated in the *ST* and *SA* of Figure 2 (left).

While searching using the *ST* or the *SA* is efficient in theory, it requires large amounts of main memory. A CST reduces the space requirements of *ST* by utilizing the compressibility of the Burrows-Wheeler transform (BWT) (Burrows and Wheeler, 1994). The BWT corresponds to a reversible permutation of the text used in data compression tools such as BZIP2 to increase the compressibility of the input. The transform is defined as

$$\text{BWT}[i] = \text{T}[SA[i] - 1 \mod n] \quad (1)$$

and is the core component of the FM-Index (Ferragina and Manzini, 2000) which is a subcomponent of a CST to provide efficient search for locating arbitrary length patterns ($m$-grams), determining occurrence frequencies etc. The key functionality provided by the FM-Index is the ability to efficiently determine the range $SA[lb, rb]$ matching a given pattern $\alpha$ described above without the need to store the *ST* or *SA* explicitly. This is achieved by iteratively processing $\alpha$ in reverse order using the BWT, which is usually referred to as *backward-search*.

The backward-search procedure utilizes the duality between the BWT and *SA* to iteratively determine $SA[lb, rb]$ for suffixes of $\alpha$. Suppose $SA[sp_j, ep_j]$ corresponds to all suffixes in $\mathcal{T}$ matching $\alpha[j, m-1]$. Range $SA[sp_{j-1}, ep_{j-1}]$ matching $\alpha[j-1, m-1]$ with $c \stackrel{\text{def}}{=} \alpha[j-1]$ can be expressed as

$$sp_{j-1} = C[c] + \text{RANK}(\text{BWT}, sp_j, c)$$
$$ep_{j-1} = C[c+1] + \text{RANK}(\text{BWT}, ep_j + 1, c) - 1$$

where $C[c]$ refers to the starting position of all suffixes prefixed by $c$ in *SA* and $\text{RANK}(\text{BWT}, sp_j, c)$ determines the number of occurrences of symbol $c$ in $\text{BWT}[0, sp_j]$.

Operation $\text{RANK}(\text{BWT}, i, c)$ (and its inverse operation $\text{SELECT}(\text{BWT},i,c)$[2]) can be performed efficiently using a *wavelet tree* (Grossi et al., 2003) representation of the BWT. A wavelet tree is a versatile, space-efficient representation of a sequence which can efficiently support a variety of operations (Navarro, 2014). The structure of the wavelet tree is derived by recursively decomposing the alphabet into subsets. At each level the alphabet is

---
[2] SELECT(BWT,i,c) returns the position of the $i$th occurrence of symbol c in BWT.

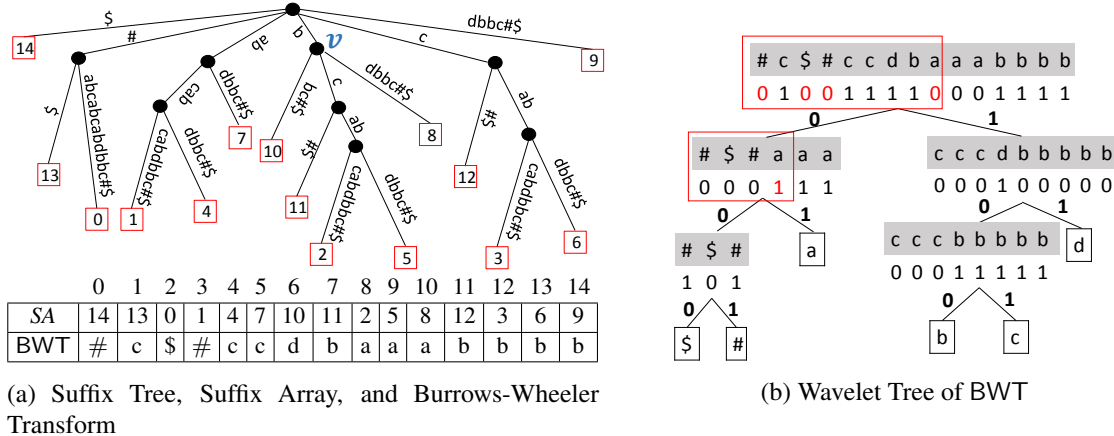

(a) Suffix Tree, Suffix Array, and Burrows-Wheeler Transform

(b) Wavelet Tree of BWT

Figure 2: (a) Character-Level Suffix Tree, Suffix Array (*SA*), and Burrows-Wheeler Transform (BWT) for "#abcab-cabdbbc#$" (as formulated in eq. 1). (b) The Wavelet Tree of BWT and the RANK(BWT, 8, a). The ordered alphabet and their code words are {$:000, #:001, a:01, b:100, c:101, d:11}, and symbols "#" and "$" are to mark sentence and file boundaries. The red bounding boxes and digits signify the path for computing RANK(BWT, 8, a).

split into two subsets based on which symbols in the sequence are assigned to the left and right child nodes respectively. Using compressed bitvector representations and Huffman codes to define the alphabet partitioning, the space usage of the wavelet tree and associated RANK structures of the BWT is bound by $H_k(T)n + o(n \log \sigma)$ bits (Grossi et al., 2003). Thus the space usage is proportional to the order $k$ entropy ($H_k(T)$) of the text.

Figure 2 (right) shows a sample wavelet tree representation. Using the wavelet tree structure, RANK over a sequence drawn from an alphabet of size $\sigma$ can be reduced to $\log \sigma$ binary RANK operations which can be answered efficiently in constant time (Jacobson, 1989). The range $SA[lb, rb]$ corresponding to a pattern $\alpha$, can be determined in $O(m \log \sigma)$ time using a wavelet tree of the BWT.

In addition to the FM-index, a CST efficiently stores the tree topology of the *ST* to emulate tree operations such efficiently (Ohlebusch et al., 2010).

### 3.2 Computing KN modified counts

Shareghi et al. (2015) showed how the requisite counts for a KN-LM, namely $c(\alpha)$, $N_{1+}(\bullet\alpha), N_{1+}(\bullet\alpha\bullet)$ and $N_{1+}(\alpha\bullet)$, can be computed directly from CST. Consider the example in Figure 2, the number of occurrences of $b$ corresponds to counting the number of leaves, size($v$), in the subtree rooted at $v$. This can be computed in $O(1)$ time by computing the size of the range $[lb, rb]$ implicitly associated with each node. The number of unique right contexts of $b$ can be determined using degree($v$) (again $O(1)$ but requires bit operations on the succinct tree representation of the *ST*). That is, $N_{1+}(b\bullet) = $ degree($v$) $= 3$.

Determining the number of left-contexts and surrounding contexts are more involved. Computing $N_{1+}(\bullet\alpha)$ relies on the BWT. Recall that BWT[$i$] corresponds to the symbol *preceding* the suffix starting at $SA[i]$. For example computing $N_{1+}(\bullet b)$ first requires finding the interval of suffixes starting with $b$ in *SA*, namely $lb = 6$ and $rb = 10$, and then counting the number of unique symbols in BWT[6, 10] = \{d, b, a, a, a\}, i.e., 3. Determining all unique symbols in BWT[$i, j$] can be performed efficiently (independently of the size of the range) using the *wavelet tree* encoding of the BWT. The set of symbols preceding pattern $\alpha$, denoted by $P(\alpha)$ can be computed in $O(|P(\alpha)| \log \sigma)$ by visiting all unique leafs in the wavelet tree which correspond to symbols in BWT[$i, j$]. This is usually referred to as the *interval-symbols* (Schnattinger et al., 2010) procedure and uses RANK operations to find the set of symbols $s \in P(\alpha)$ and corresponding ranges for $s\alpha$ in *SA*. In the above example, identifying the *SA* range of $ab$ requires to find the $lb, rb$ in the *SA* for suffixes starting with $a$ (*SA* [3,5]) and then adjusting the bounds to cover only the suffixes starting with $ab$. This last step is done via computing the rank of three $a$ symbols in BWT[8,10] using the wavelet tree, see Figure 2 (right) for RANK(BWT, a, 8). As illustrated, answering RANK(BWT, 8, a) corresponds to processing the first digit of the code word at the root level, which translates into RANK($WT_{root}$, 8, 0) = 4, followed by a RANK($WT_1$, 4, 1) = 1 on the left branch. Once the ranks are computed $lb, rb$ are refined accordingly to *SA* [3+ (1 - 1), 3+ (3 - 1)]. Finally, for $N_{1+}(\bullet \alpha \bullet)$ all patterns which can follow $\alpha$ are enumerated, and for each of these extended patterns, the number of preceding symbols is computed using *interval-symbols*. This proved to be the most expensive operation in their approach.

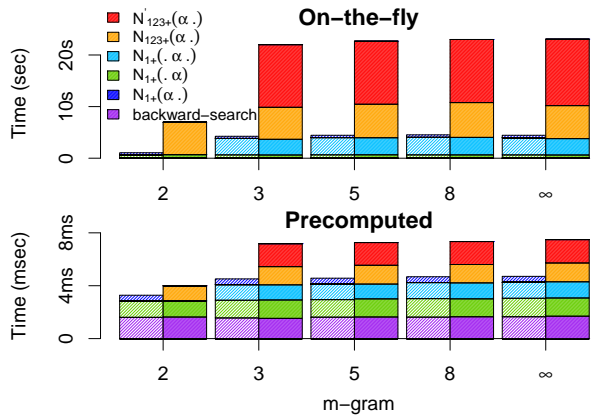

Figure 3: Time breakdown for querying average per-sentence, shown without runtime precomputation of expensive contextual counts (above) vs. with precomputation (below). The left and right bar in each group denote KN and MKN, respectively. Trained on the German portion of the Europarl corpus and tested over the first $10K$ sentences from the News Commentary corpus.

Given these quantities, Shareghi et al. (2015) show how $m$-gram probabilities can be computed on demand using an iterative algorithm to search for matching nodes in the suffix tree for the required $k$-gram ($k \leq m$) patterns in the numerator and denominator of KN recursive equations, which are then used to compute the probabilities. We refer the reader to Shareghi et al. (2015) for further details. Overall their approach showed promise, in that it allowed for unlimited order KN-LMs to be evaluated with a modest memory footprint, however it was many orders of magnitude slower for smaller $m$ than leading LM toolkits.

To illustrate the cost of querying, see Figure 3 (top) which shows per-sentence query time for KN,

---

**Algorithm 1** $N_{\{1,2,3+\}}(\alpha \bullet)$ or $N'_{\{1,2,3+\}}(\alpha \bullet)$

1: **function** N123PFRONT($t, v, \alpha$, is-prime)
        ▷ $t$ is a CST, $v$ is the node matching pattern $\alpha$
2:     $N_1, N_2, N_{3+} \leftarrow 0$
3:     **for** $u \leftarrow$ children($v$) **do**
4:         **if** is-prime **then**
5:             $f \leftarrow$ interval-symbols($t$, [lb($u$), rb($u$)])
6:         **else**
7:             $f \leftarrow$ size($u$)
8:         **if** $1 \leq f \leq 2$ **then**
9:             $N_f \leftarrow N_f + 1$
10:    $N_{3+} \leftarrow$ degree($v$) $- N_1 - N_2$
11:    **return** $N_1, N_2, N_{3+}$

---

based on the approach of Shareghi et al. (2015) (also shown is MKN, through an extension of their method as described in §4). It is clear that the runtimes for KN is much too slow for practical use – about 5 seconds per sentence, with the majority of this time spent computing $N_{1+}(\bullet \alpha \bullet)$. Clearly optimisation is warranted, and the gains from doing so are spectacular (see Figure 3 bottom, which uses the precomputation method as described in §4.2).

## 4 Extending to MKN

### 4.1 Computing MKN modified counts

A central requirement for extending Shareghi et al. (2015) to support MKN are algorithms for computing $N_{\{1,2,3+\}}(\alpha \bullet)$ and $N'_{\{1,2,3+\}}(\alpha \bullet)$, which we now expound upon. Algorithm 1 computes both of these quantities, taking as input a CST $t$, a node $v$ matching the pattern $\alpha$, the pattern and a flag is-prime, denoting which of the $N$ and $N'$ variants is required. This method enumerates the children of the node (line 3) and calculates either the frequency of each child (line 7) or the modified count $N_{1+}(\bullet \alpha$ x), for each child $u$ where x is the first symbol on the edge $vu$ (line 5). Lines 8 and 9 accumulate the number of these values equal to one or two, and finally in line 10, $N_{3+}$ is computed by the difference between $N_{1+}(\alpha \bullet) = $ degree($v$) and the already counted events $N_1 + N_2$.

For example, computing $N_{\{1,2,3+\}}(b \bullet)$ in Figure 2 corresponds to enumerating over its three children. Two of $v$'s children are leaf nodes $\{10, 8\}$, and one child has three leaf descendants $\{11, 2, 5\}$, hence $N_1$ and $N_2$ are 2 and 0 respectively, and $N_{3+}$ is 1. Further, consider computing $N'_{\{1,2,3+\}}(b \bullet)$ in

Figure 2, which again enumerates over child nodes (whose path labels start with symbols $b, c$ and $d$) and computes the number of preceding symbols for the extended patterns.[3] Accordingly $N'_1(b\, \bullet) = 2$, $N'_2(b\, \bullet) = 1$ and $N'_{3+}(b\, \bullet) = 0$.

While roughly similar in approach, computing $N'_{\{1,2,3+\}}(\alpha\, \bullet)$ is in practice slower than $N_{\{1,2,3+\}}(\alpha\, \bullet)$ since it requires calling *interval-symbols* (line 7) instead of calling the constant time *size* operation (line 5). This gives rise to a time complexity of $O(d|P(\alpha)|\log \sigma)$ for $N'_{\{1,2,3+\}}(\alpha\, \bullet)$ where $d$ is the number of children of $v$.

As illustrated in Figure 3 (top), the modified counts (§2) combined are responsible for 99% of the query time. Moreover the already expensive runtime of KN is considerably worsened in MKN due to the additional counts required. These facts motivate optimisation, which we achieve by precomputing values, resulting in a 2500× speed up in query time as shown in Figure 3 (bottom).

### 4.2 Efficient Precomputation

Language modelling toolkits such as KenLM and SRILM precompute real valued probabilities and backoff-weights at training time, such that querying becomes largely a problem of retrieval. We might consider taking a similar route in optimising our language model, however we would face the problem that floating point numbers cannot be compressed very effectively. Even with quantisation, which can have a detrimental effect on modelling perplexity, we would not expect good compression and thus this technique would limit the scaling potential of our approach.

For these reasons, instead we store the most expensive count data, targeting those counts which have the greatest effect on runtime (see Figure 3 top). We expect these integer values to compress well: as highlighted by Figure 4 most counts will have low values, and accordingly a variable byte compression scheme will be able to realise high compression rates. Removing the need for computing these values at query time leaves only pattern search and a few floating point operations in order to compute language model probabilities (see §4.3) which can be done cheaply.

---
[3]That is $N_{1+}(\bullet\, bb) = 1$, $N_{1+}(\bullet\, bc) = 2$, $N_{1+}(\bullet\, bd) = 1$.

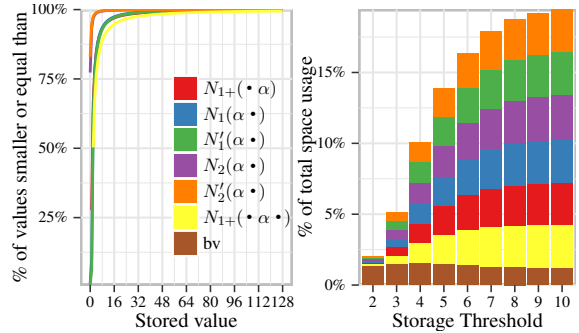

Figure 4: Left: Distribution of values prestored for Europarl German; Right: Space usage of prestored values relative to total index size for Europarl German for different storage thresholds ($\hat{m}$).

Our first consideration is how to structure the cache. Given that each precomputed value is computed using a CST node, $v$, (with the pattern as its path-label), we structure the cache as a mapping between unique node identifiers $\text{id}(v)$ and the precomputed values.[4] Next we consider which values to cache: while it is possible to precompute values for every node in the CST, many nodes are unlikely to be accessed at query time. Moreover, these rare patterns are likely to be cheap to process using the on-the-fly methods, given they occur in few contexts. Consequently precomputing these values will bring minimal speed benefits, while still incurring a memory cost. For this reason we precompute the values only for nodes corresponding to $k$-grams up to length $\hat{m}$ (for our word-level experiments $\hat{m} = 10$), which are most likely to be accessed at runtime.[5]

The precomputation method is outlined in Algorithm 2, showing how a compressed cache is created for the quantities $x \in \{N_{1+}(\bullet\, \alpha),\ N_{1+}(\bullet\, \alpha\, \bullet),\ N_{12}(\alpha\, \bullet),\ N'_{12}(\alpha\, \bullet)\}$. The algorithm visits the suffix tree nodes in depth-first-search (DFS) order, and selects a subset of nodes for precomputation (line 7), such that the remaining nodes are either rare or trivial to handle

---
[4]Each node can uniquely be identified by the order which it is visited in a DFS traversal of the suffix tree. This corresponds to the RANK of the opening parenthesis of the node in the balanced parenthesis representation of the tree topology of the CST which can be determined in $O(1)$ time (Ohlebusch et al., 2010).

[5]We did not test other selection criteria. Other methods may be more effective, such as selecting nodes for precomputation based on the frequency of their corresponding patterns in the training set.

**Algorithm 2** Precomputing expensive counts $N_{\{1,2\}}(\alpha \bullet), N_{1+}(\bullet \alpha \bullet), N_{1+}(\bullet \alpha), N'_{\{1,2\}}(\alpha \bullet)$.

1: **function** PRECOMPUTE($t, \hat{m}$)
2:     $bv_l \leftarrow 0 \quad \forall l \in [0, \text{nodes}(t) - 1]$
3:     $i_l^{(x)} \leftarrow 0 \quad \forall l \in [0, \text{nodes}(t) - 1], x \in \text{count types}$
4:     $j \leftarrow 0$
5:     **for** $v \leftarrow \text{descendants}(\text{root}(t))$ **do**     ▷ DFS
6:         $d \leftarrow \text{string-depth}(v)$
7:         **if** not is-leaf($v$) $\wedge\ d \leq \hat{m}$ **then**
8:             $l \leftarrow \text{id}(v)$     ▷ unique id
9:             $bv_l \leftarrow 1$
10:            Call N1PFRONTBACK1($t, v, \cdot$)
11:            Call N123PFRONT($t, v, \cdot, 0$)
12:            Call N123PFRONT($t, v, \cdot, 1$)
13:            $i_j^{(x)} \leftarrow$ counts from above, for each output $x$
14:            $j \leftarrow j + 1$
15:     $bv_{rrr} \leftarrow \text{compress-rrr}(bv)$
16:     $i \leftarrow \text{compress-dac}(\{i^{(x)}\ \forall x\})$
17:     write-to-disk($bv_{rrr}, i$)

on-the-fly (i.e., leaf nodes). A node included in the cache is marked by storing a 1 in the bit vector $bv$ (lines 8-9) at index $l$, where $l$ is the node identifier. For each selected node we precompute the expensive counts in lines 10-12,

$N_{1+}(\bullet \alpha \bullet), N_{1+}(\bullet \alpha)$ via[6] N1PFRONTBACK1($t, v, \cdot$),

$N_{\{1,2\}}(\alpha \bullet)$ via N123PFRONT($t, v, \cdot, 0$),

$N'_{\{1,2\}}(\alpha \bullet)$ via N123PFRONT($t, v, \cdot, 1$),

which are stored into integer vectors $i^{(x)}$ for each count type $x$ (line 13). The integer vectors are streamed to disk and then compressed (lines 15-17) in order to limit memory usage.

The final steps in lines 15 and 16 compress the integer and bit-vectors. The integer vectors $i^{(x)}$ are compressed using a variable length encoding, namely Directly Addressable Variable-Length Codes (DAC; Brisaboa et al. (2009)) which allows for efficient storage of integers while providing efficient random access. As the overwhelming majority of our precomputed values are small (see Figure 4 left), this gives rise to a dramatic compression rate of only $\approx 5.2$ bits per integer. The bit vector $bv$ of size $O(n)$ where $n$ is the number of nodes in the suffix tree, is compressed using the scheme of Raman et al. (2002) which supports constant time rank operation over very large bit vectors.

This encoding allows for efficient retrieval of the precomputed counts at query time. The compressed vectors are loaded into memory and when an expensive count is required for node $v$, the precomputed quantities can be fetched in constant time via LOOKUP($v, bv, i^{(x)}) = i_{\text{RANK}(bv, \text{id}(v), 1)}^{(x)}$. We use RANK to determine the number of 1s preceding $v$'s position in the bit vector $bv$. This corresponds to $v$'s index in the compressed integer vectors $i^{(x)}$, from which its precomputed count can be fetched. This strategy only applies for precomputed nodes; for other nodes, the values are computed on-the-fly.

Figure 3 compares the query time breakdown for on-the-fly count computation (top) versus precomputation (bottom), for both KN and MKN and with different Markov orders, $m$. Note that query speed improves dramatically, by a factor of about $2500\times$, for precomputed cases. This improvement comes at a modest cost in construction space. Precomputing for CST nodes with $m \leq 10$ resulted in 20% of the nodes being selected for precomputation. The space used by the precomputed values accounts for 20% of the total space usage (see Figure 4 right). Index construction time increased by 70%.

### 4.3 Computing MKN Probability

Having established a means of computing the requisite counts for MKN and an efficient precomputation strategy, we now turn to the algorithm for computing the language model probability. This is presented in Algorithm 3, which is based on Shareghi et al. (2015)'s single CST approach for computing the KN probability (reported in their paper as Algorithm 4.) Similar to their method, our approach implements the recursive $m$-gram probability formulation as an iterative loop (here using MKN). The core of the algorithm are the two nodes $v^{\text{full}}$ and $v$ which correspond to nodes matching the full $k$-gram and its $(k-1)$-gram context, respectively.

Although similar to Shareghi et al. (2015)'s method, which also features a similar right-to-left pattern lookup, in addition we optimise the computation of a full sentence probability by sliding a window of width $m$ over the sequence from left-to-right, adding one new word at a time.[7] This allows for the

---

[6] The function N1PFRONTBACK1 is defined as Algorithm 5 in Shareghi et al. (2015).

[7] Pauls and Klein (2011) propose a similar algorithm for trie-based LMs.

**Algorithm 3** MKN probability $P(w_i | w_{i-(m-1)}^{i-1})$

1: **function** PROBMKN($t, w_{i-m+1}^i, m, [v_k]_{k=0}^{m-1}$)
2:     **Assumption:** $v_k$ is the matching node for $w_{i-k}^{i-1}$
3:     $v_0^{\text{full}} \leftarrow \text{root}(t)$     ▷ tracks match for $w_{i-k}^i$
4:     $p \leftarrow 1/|\sigma|$
5:     **for** $k \leftarrow 1$ to $m$ **do**
6:         **if** $v_{k-1}$ does not match **then**
7:             break out of loop
8:         $v_k^{\text{full}} \leftarrow \text{back-search}([\text{lb}(v_{k-1}^{\text{full}}), \text{rb}(v_{k-1}^{\text{full}})], w_{i-k+1})$
9:         $\mathbb{D}^k(1), \mathbb{D}^k(2), \mathbb{D}^k(3+) \leftarrow$ discounts for $k$-grams
10:         **if** $k = m$ **then**
11:             $c \leftarrow \text{size}(v_k^{\text{full}})$
12:             $d \leftarrow \text{size}(v_{k-1})$
13:             $N_{1,2,3+} \leftarrow \text{N123PFRONT}(t, v_{k-1}, w_{i-k+1}^{i-1}, 0)$
14:         **else**
15:             $c \leftarrow \text{N1PBACK1}(t, v_k^{\text{full}}, w_{i-k+1}^{i-1})$
16:             $d \leftarrow \text{N1PFRONTBACK1}(t, v_{k-1}, w_{i-k+1}^{i-1})$
17:             $N_{1,2,3+} \leftarrow \text{N123PFRONT}(t, v_{k-1}, w_{i-k+1}^{i-1}, 1)$
18:         **if** $1 \le c \le 2$ **then**
19:             $c \leftarrow c - \mathbb{D}^k(c)$
20:         **else**
21:             $c \leftarrow c - \mathbb{D}^k(3+)$
22:         $\gamma \leftarrow \mathbb{D}^k(1)N_1 + \mathbb{D}^k(2)N_2 + \mathbb{D}^k(3+)N_{3+}$
23:         $p \leftarrow \frac{1}{d}(c + \gamma p)$
24:     **return** $\left(p, [v_k^{\text{full}}]_{k=0}^{m-1}\right)$

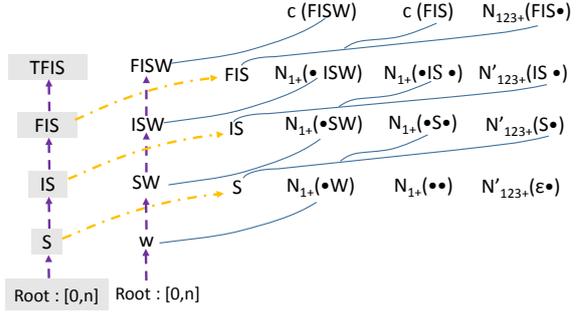

Figure 5: Example MKN probability computation for a 4-gram LM applied to *"The Force is strong with"* (each word abbreviated to its first character), showing in the two left columns the suffix matches required for the 4-gram FISW and elements which can be reused from previous 4-gram computation (gray shading), TFIS. Elements on the right denote the count and occurrence statistics derived from the suffix matches, as linked by blue lines.

re-use of nodes in one window matching the full $k$-grams, $v^{\text{full}}$, as the nodes matching the context in the subsequent window, denoted $v$.

For example, in the sentence *"The Force is strong with this one."*, computing the 4-gram probability of *"The Force is strong"* requires matches into the CST for *"strong"*, *"is strong"*, etc. As illustrated in Table 1, for the next 4-gram resulting from sliding the window to include *"with"*, the denominator terms require exactly these nodes, see Figure 5. Practically, this is achieved by storing the matching $v^{\text{full}}$ nodes computed in line 8, and passing this vector as the input argument $[v_k]_{k=0}^{m-1}$ to the next call to PROBMKN (line 1). This saves half the calls to backward-search, which, as shown in Figure 3, represent a significant fraction of the querying cost, resulting in a 30% improvement in query runtime.

The algorithm starts by considering the unigram probability, and grows the context to its left by one word at a time until the $m$-gram is fully covered (line 5). This best suits the use of backward-search in a CST, which proceeds from right-to-left over the search pattern. At each stage the search for $v_k^{\text{full}}$ uses the span from the previous match, $v_{k-1}^{\text{full}}$, along with the BWT to efficiently locate the matching node. Once the nodes matching the full sequence and its context are retrieved, the procedure is fairly straightforward: the discounts are loaded on line 9 and applied in lines 18-21, while the numerator, denominator and smoothing quantities as required for computing $P$ and $\bar{P}$ are calculated in lines 10-13 and 15-17, respectively.[8] Note that the calls for functions N123PFRONT, N1PBACK1, N1PFRONTBACK1 are avoided if the corresponding node is amongst the selected nodes in the precomputation step; instead the LOOKUP function is called. Finally, the smoothing weight $\gamma$ is computed in line 22 and the conditional probability computed on line 23. The loop terminates when we reach the length limit $k = m$ or we cannot match the context, i.e., $w_{i-k}^{i-1}$ is not in the training corpus, in which case the probability value $p$ for the longest match is returned.

We now turn to the discount parameters, $\mathbb{D}^k(j), k \le m, j \in 1, 2, 3+$, which are function of the corpus statistics as outlined in Figure 1. While these could be computed based on raw $m$-gram statistics, this approach is very inefficient for large $m \ge 5$; instead these values can be computed efficiently from the compressed data structures. Algorithm 4 outlines how the $\mathbb{D}^k(i)$ values can be com-

---
[8]N1PBACK1 and N1PFRONTBACK1 are defined in Sharehi et al. (2015); see also §3 for an overview.

**Algorithm 4** Compute discounts

1: **function** COMPUTEDISCOUNTS($t, \bar{m}, bv', SA$)
2:     $n_i(k) \leftarrow 0, \bar{n}_i(k) \leftarrow 0 \quad \forall i \in [1,4], k \in [1,\bar{m}]$
3:     $N_{1+}(\bullet \bullet) \leftarrow 0$
4:     **for** $v \leftarrow$ descendants(root($t$)) **do**     ▷ DFS
5:         $d_P \leftarrow$ string-depth(parent($v$))
6:         $d \leftarrow$ string-depth($v$)
7:         $d_S \leftarrow$ depth-next-sentinel($SA, bv', \text{lb}(v)$)
8:         $i \leftarrow$ size($v$)     ▷ frequency
9:         $c \leftarrow$ interval-symbols($t, [\text{lb}(v), \text{rb}(v)]$)   ▷ left occ.
10:        **for** $k \leftarrow d_P + 1$ to $\min(d, \bar{m}, d_S - 1)$ **do**
11:            **if** $k = 2$ **then**
12:                $N_{1+}(\bullet \bullet) \leftarrow N_{1+}(\bullet \bullet) + 1$
13:            **if** $1 \leq i \leq 4$ **then**
14:                $n_i(k) \leftarrow n_i(k) + 1$
15:            **if** $1 \leq c \leq 4$ **then**
16:                $\bar{n}_c(k) \leftarrow \bar{n}_c(k) + 1$
17:     $\mathbb{D}^k(i) \leftarrow$ computed using formula in Figure 1
18:     **return** $\mathbb{D}^k(i), k \in [1, \bar{m}], i \in \{1, 2, 3+\}$

puted directly from the CST. This method iterates over the nodes in the suffix tree, and for each node considers the $k$-grams encoded in the edge label, where each $k$-gram is taken to start at the root node (to avoid duplicate counting, we consider $k$-grams only contained on the given edge but not in the parent edges, i.e., by bounding $k$ based on the string depth of the parent and current nodes, $d_P \leq k \leq d$). For each $k$-gram we record its count, $i$ (line 8), and the number of unique symbols to the left, $c$ (line 9), which are accumulated in an array for each $k$-gram size for values between 1 and 4 (lines 13-14 and 15-16, respectively). We also record the number of unique bigrams by incrementing a counter during the traversal (lines 11-12).

Special care is required to exclude edge labels that span sentence boundaries, by detecting special sentinel symbols (line 8) that separate each sentence or conclude the corpus. This check could be done by repeatedly calling edge($v, k$) to find the $k^{th}$ symbol on the given edge to check for sentinels, however this is a slow operation as it requires multiple backward search calls. Instead we precalculate a bit vector, $bv'$, of size equal to the number of tokens in the corpus, $n$, in which sentinel locations in the text are marked by 1 bits. Coupled with this, we use the suffix array $SA$, such that

depth-next-sentinel($SA, bv', \ell$) =
    SELECT($bv'$, RANK($bv', SA_\ell, 1$) + 1, 1) − $SA_\ell$,

where $SA_\ell$ returns the offset into the text for index $\ell$, and the $SA$ is stored uncompressed to avoid the expensive cost of recovering these values.[9] This function can be understood as finding the first occurrence of the pattern in the text (using $SA_\ell$) then finding the location of the next 1 in the bit vector, using constant time RANK and SELECT operations. This locates the next sentinel in the text, after which it computes the distance to the start of the pattern. Using this method in place of explicit edge calls improved the training runtime substantially up to 41×.

We precompute the discount values for $k \leq \bar{m}$-grams. For querying with $m > \bar{m}$ (including $\infty$) we reuse the discounts for the largest $\bar{m}$-grams.[10]

## 5 Experiments

To evaluate our approach we measure memory and time usage, along with the predictive perplexity score of word-level LMs on a number of different corpora varying in size and domain. For all of our word-level LMs, we use $\bar{m}, \hat{m} \leq 10$. We also demonstrate the positive impact of increasing the set limit on $\bar{m}, \hat{m}$ from 10 to 50 on improving character-level LM perplexity. The SDSL library (Gog et al., 2014) is used to implement our data structures. The benchmarking experiments were run on a single core of a Intel Xeon E5-2687 v3 3.10GHz server with 500GiB of RAM.

In our word-level experiments, we use the German subset of the Europarl (Koehn, 2005) as a small corpus, which is 382 MiB in size measuring the raw uncompressed text. We also evaluate on much larger corpora, training on 32GiB subsets of the deduplicated English, Spanish, German, and French Common Crawl corpus (Buck et al., 2014). As test sets, we used newstest-2014 for all languages except Spanish, for which we used newstest-2013.[11] In our benchmarking experiments we used the bottom 1M sentences (not used in training) of German Common Crawl corpus. We used the preprocessing script of Buck et al. (2014), then removed sentences with $\leq 2$ words, and replaced rare words[12] $c \leq 9$ in the training data with a special token. In our character-level experiments, we used the training and test data of the benchmark 1-billion-words corpus (Chelba et al., 2013).

---

[9] Although the $SA$ can be very large, we need not store it in memory. The DFS traversal in Algorithm 4 (lines 4–16) means that the calls to $SA_\ell$ occur in increasing order of $\ell$. Hence, we use on-disk storage for the $SA$ with a small memory mapped buffer, thereby incurring a negligible memory overhead.

[10] It is possible to compute the discounts for all patterns of the text using our algorithm with the complexity linear in the length of the text. However, the discounts appear to converge by pattern length $\bar{m} = 10$. This limit also helps to avoid problems of wild fluctuations in discounts for very long patterns arising from noise for low count events.

[11] http://www.statmt.org/wmt{13,14}/test.tgz

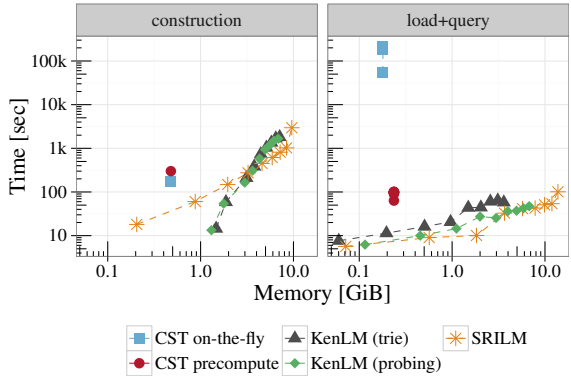

Figure 6: Memory consumption and *total* runtime for the CST with and without precomputation, KenLM (trie), and SRILM (default) with $m \in [2, 10]$, while we also include $m = \infty$ for CST methods. Trained on the Europarl German corpus and tested over the bottom 1M sentences from German Common Crawl corpus.

| size (M) | | | perplexity | | | |
|---|---|---|---|---|---|---|
| tokens | $m=2$ | $m=3$ | $m=5$ | $m=7$ | $m=10$ | $m=\infty$ |
| EN 6470 | 321.6 | 183.8 | 154.3 | 152.7 | 152.5 | 152.3 |
| ES 6276 | 231.3 | 133.2 | 111.7 | 109.7 | 109.3 | 109.2 |
| FR 6100 | 215.8 | 109.2 | 85.2 | 83.1 | 82.6 | 82.4 |
| DE 5540 | 588.3 | 336.6 | 292.8 | 288.1 | 287.8 | 287.8 |

Table 2: Perplexities on English, French, German newstests 2014, and Spanish newstest 2013 when trained on 32GiB chunks of English, Spanish, French, and German Common Crawl corpus.

**Small data: German Europarl** First, we compare the time and memory consumption of both the SRILM and KenLM toolkits, and the CST on the small German corpus. Figure 6 shows the memory usage for construction and querying for CST-based methods w/o precomputation is independent of $m$, but they grow substantially with $m$ for the SRILM and KenLM benchmarks. To make our results comparable to those reported in (Shareghi et al., 2015) for query time measurements we reported the loading and query time combined. The construction cost is modest, requiring less memory than the benchmark systems for $m \geq 3$, and running in a similar time[13] (despite our method supporting queries of unlimited size). Precomputation adds to the construction time, which rose from 173 to 299 seconds, but yielded speed improvements of several orders of magnitude for querying (218k to 98 seconds for 10-gram). In querying, the CST-precompute method is 2-4× slower than both SRILM and KenLM for large $m \geq 5$, with the exception of $m = 10$ where it outperforms SRILM. A substantial fraction of the query time is loading the structures from disk; when this cost is excluded, our approach is between 8-13× slower than the benchmark toolkits. Note that perplexity computed by the CST closely matched KenLM (differences $\leq 0.1$).

**Big Data: Common Crawl** Table 2 reports the perplexity results for training on 32GiB subsets of the English, Spanish, French, and German Common Crawl corpus. Note that with such large datasets, perplexity improves with increasing $m$, with substantial gains available moving above the widely used $m = 5$. This highlights the importance of our approach being independent from $m$, in that we can evaluate for any $m$, including $\infty$, at low cost.

**Heterogeneous Data** To illustrate the effects of domain shift, corpus size and language model capacity on modelling accuracy, we now evaluate the system using a variety of different training corpora. Table 3 reports the perplexity for German when training over datasets ranging from the small Europarl up to 32GiB of the Common Crawl corpus. Note that the test set is from the same domain as the News Crawl, which explains the vast difference in perplexities. The domain effect is strong enough to eliminate the impact of using much larger corpora, compare 10-gram perplexities for training on the smaller News Crawl 2007 corpus versus Europarl. However 'big data' is still useful: in all cases the perplexity improves as we provide more data from the same source. Moreover, the magnitude of the gain in perplexity when increasing $m$ is influenced by the data size: with more training data higher order $m$-grams provide richer models; therefore, the scalability of our method to large datasets is crucially important.

---

[12]Running with the full vocabulary increased the memory requirement by 40% for construction and 5% for querying with our model, and 10% and 30%, resp. for KenLM. Construction times for both approaches were 15% slower, but query runtime was 20% slower for our model versus 80% for KenLM.

[13]For all timings reported in the paper we manually flushed the system cache between each operation (both for construction and querying) to remove the effect of caching on runtime. To query KenLM, we used the speed optimised *populate* method. (We also compare the memory optimised *lazy* method in Figure 7.) To train and query SRILM we used the *default* method which is optimised for speed, but had slightly worse memory usage than the *compact* method.

|  | size (M) |  | perplexity |  |  |
|---|---|---|---|---|---|
| Training | tokens | sents | $m=3$ | $m=5$ | $m=10$ |
| Europarl | 55 | 2.2 | 1004.8 | 973.3 | 971.4 |
| Commentary | 6 | 0.2 | 941.8 | 928.6 | 928.1 |
| NCrawl2007 | 37 | 2.0 | 514.8 | 493.5 | 488.9 |
| NCrawl2008 | 126 | 6.8 | 427.7 | 404.8 | 400.0 |
| NCrawl2009 | 119 | 6.5 | 433.4 | 408.9 | 404.7 |
| NCrawl2010 | 54 | 3.0 | 472.7 | 450.9 | 446.8 |
| NCrawl2011 | 298 | 16.3 | 362.6 | 335.9 | 327.9 |
| NCrawl2012 | 377 | 20.9 | 343.9 | 315.6 | 307.3 |
| NCrawl2013 | 641 | 35.1 | 268.9 | 229.8 | 225.6 |
| NCrawl2014 | 845 | 46.3 | 247.6 | 195.2 | 189.3 |
| All combined | 2560 | 139.3 | 211.8 | 158.9 | 151.5 |
| CCrawl1G | 173 | 14.2 | 542.8 | 515.5 | 512.7 |
| CCrawl2G | 346 | 28.5 | 493.2 | 462.3 | 459.2 |
| CCrawl4G | 692 | 56.2 | 446.5 | 412.2 | 408.8 |
| CCrawl8G | 1390 | 110.2 | 402.1 | 364.4 | 360.9 |
| CCrawl16G | 2770 | 216.7 | 364.9 | 323.9 | 319.6 |
| CCrawl32G | 5540 | 426.6 | 336.6 | 292.8 | 287.8 |

Table 3: Perplexity of German newstest 2014 with different datasets (Europarl, News-Commentary, NewsCrawl 2007-2014, CommonCrawl 1-32 GiB chunks) and $m$.

**Benchmarking against KenLM** Next we compare our model against the state-of-the-art method, KenLM trie. The perplexity difference between CST and KenLM was less than 0.003 in all experiments.

*Construction Cost.* Figure 7a) compares the peak memory usage of our CST models and KenLM. KenLM is given a target memory usage of the peak usage of our CST models.[14] The construction phase for the CST required more time for lower order models (see Figure 7c) but was comparable for larger $m$, roughly matching KenLM for $m = 10$.[15] For the 32GiB dataset, the CST model took 14 hours to build, compared to KenLM's 13.5 and 4 hours for the 10-gram and 5-gram models, respectively.

---

[14]Using the memory budget option, *-S*. Note that KenLM often used more memory than specified. Allowing KenLM use of 80% of the available RAM reduced training time by a factor of between 2 and 4.

*Query Cost.* As shown in Figure 7b, the memory requirements for querying with the CST method were consistently lower than KenLM for $m \geq 4$: for $m = 10$ the memory consumption of KenLM was 277GiB compared to our 27GiB, a 10× improvement. This closely matches the file sizes of the stored models on disk. Figure 7d reports the query runtimes, showing that KenLM grows substantially slower with increasing dataset size and increasing language model order. In contrast, the runtime of our CST approach is much less affected by data size or model order. Our approach is faster than KenLM with the memory optimised *lazy* option for $m \geq 3$, often by several orders of magnitude. For the faster KenLM *populate*, our model is still highly competitive, growing to 4× faster for the largest data size.[16] The loading time is still a significant part of the runtime; without this cost, our model is 5× slower than KenLM *populate* for $m = 10$ on the largest dataset. Running our model with $m = \infty$ on the largest data size did not change the memory usage and only had a minor effect on runtime, taking 645s.

**Character-level modelling** To demonstrate the full potential of our approach, we now consider character based language modelling, evaluated on the large benchmark 1-billion-words language modelling corpus, a 3.9GiB (training) dataset with 768M words and 4 billion characters.[17] Table 4 shows the test perplexity results for our models, using the full training vocabulary. Note that perplexity improves with $m$ for the character based model, but plateaus at $m = 10$ for the word based model; one reason for this is the limited discount computation, $\bar{m} \leq 10$,

---

[15]The CST method uses a single thread for construction, while KenLM uses several threads. Most stages of construction for our method could be easily parallelised.

[16]KenLM benefits significantly from caching which can occur between runs or as more queries are issued (from $m$-gram repetition in our large 1 million sentence test set), whereas the CST approach does not benefit noticeably (as it does not incorporate any caching functionality).

[17]http://www.statmt.org/lm-benchmark/

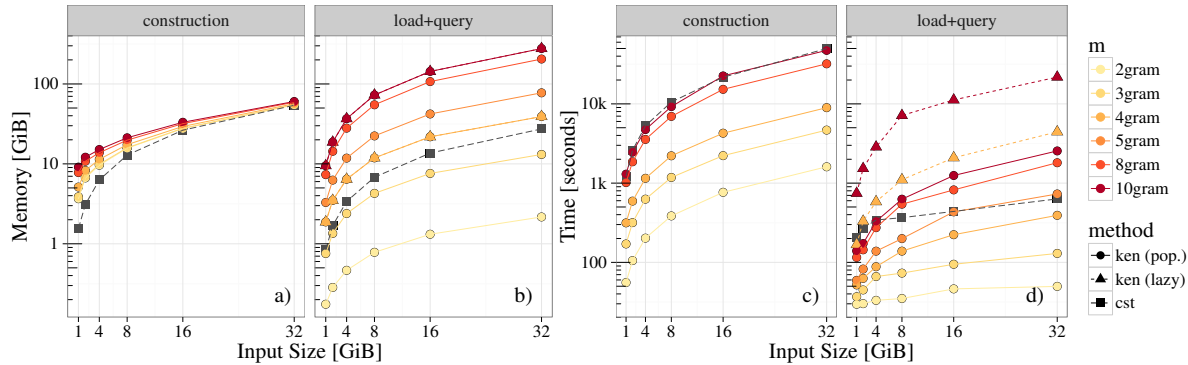

Figure 7: Memory and runtime statistics for CST and KenLM for construction and querying with different amounts of German Common Crawl training data and different Markov orders, $m$. We compare the query runtimes against the optimised version of KenLM for memory (*lazy*) and speed (*populate*). For clarity, in the figure we only show CST numbers for $m = 10$; the results for other settings of $m$ are very similar. KenLM was trained to match the construction memory requirements of the CST-precompute method.

| unit | time (s) | mem (GiB) | $m = 5$ | $m = 10$ | $m = 20$ | $m = \infty$ |
|---|---|---|---|---|---|---|
| word | 8164 | 6.29 | 73.45 | 68.66 | 68.76 | 68.80 |
| character | 17 935 | 18.58 | 3.93 | 2.69 | 2.37 | 2.33 |

Table 4: Perplexity results for the 1 billion word benchmark corpus, showing word based and character based MKN models, for different $m$. Timings and peak memory usage are reported for construction. The word model computed discounts and precomputed counts up to $\bar{m}, \hat{m} = 10$, while the character model used thresholds $\bar{m}, \hat{m} = 50$. Timings measured on a single core.

for the word model, which may not be a good parameterisation for $m > \bar{m}$.

Despite the character based model (implicitly) having a massive parameter space, estimating this model was tractable with our approach: the construction time was a modest 5 hours (and 2.3 hours for the word based model.) For the same dataset, Chelba et al. (2013) report that training a MKN 5-gram model took 3 hours using a cluster of 100 CPUs; our algorithm is faster than this, despite only using a single CPU core.[18] Queries were also fast: 0.72-0.87ms and 15ms per sentence for word and character based models, respectively.

## 6 Conclusions

We proposed a language model based on *compressed suffix trees*, a representation that is highly compact and can be easily held in memory, while supporting queries needed in computing language model probabilities on the fly. We presented several optimisations to accelerate this process, with only a modest increase in construction time and memory usage, yet improving query runtimes up to $2500\times$. In benchmarking against the state-of-the-art KenLM package on large corpora, our method has superior memory usage and highly competitive runtimes for both querying and training. Our approach allows easy experimentation with high order language models, and our results provide evidence that such high orders most useful when using large training sets.

We posit that further perplexity gains can be realised using richer smoothing techniques, such as a non-parametric Bayesian prior (Teh, 2006; Wood et al., 2011). Our ongoing work will explore this avenue, as well as integrating our language model into the Moses machine translation system, and improving the querying time by caching the lower order probabilities (e.g., $m < 4$) which we believe can improve query time substantially while maintaining a modest memory footprint.

## Acknowledgements

This research was supported by the Australian Research Council (FT130101105), National ICT Australia (NICTA) and a Google Faculty Research Award.

---

[18] Chelba et al. (2013) report a better perplexity of 67.6, but they pruned the training vocabulary, whereas we did not. Also we use a stringent treatment of OOV, following Heafield (2013).